\documentclass[10pt]{article}

\usepackage{amsmath}
\usepackage{amssymb}

\usepackage{graphicx}

\usepackage{cite}

\usepackage{color}

\usepackage[labelfont=bf,labelsep=period,justification=raggedright]{caption}
\def\DI{{\mathrm {DI}}}
\makeatletter
\renewcommand{\@biblabel}[1]{\quad#1.}
\makeatother

\date{}

\begin{document}

\begin{flushleft}
{\Large \textbf{Shrinkage Optimized Directed Information using
Pictorial Structures for Action Recognition} }
\\
Xu Chen$^{1,\ast}$, Alfred Hero$^{1}$, Silvio Savarese$^{1}$
\\
\bf{1}  Dept of EECS, University of Michigan at Ann Arbor, Ann
Arbor, MI, USA
\\

$\ast$ E-mail: Corresponding xhen@umich.edu
\end{flushleft}

\section*{Abstract}
In this paper, we propose a novel action recognition framework. The
method uses pictorial structures and shrinkage optimized directed
information assessment (SODA) coupled with Markov Random Fields
called SODA+MRF to model the directional temporal dependency and
bidirectional spatial dependency. As a variant of mutual
information, directional information captures the directional
information flow and temporal structure of video sequences across
frames. Meanwhile, within each frame, Markov random fields are
utilized to model the spatial relations among different parts of a
human body and the body parts of different people. The proposed
SODA+MRF model is robust to view point transformations and detect
complex interactions accurately. We compare the proposed method
against several baseline methods to highlight the effectiveness of
the SODA+MRF model. We demonstrate that our algorithm has superior
action recognition performance on the UCF action recognition
dataset, the Olympic sports dataset and the collective activity
dataset over several state-of-the-art methods.

\section{Introduction}
The ability to accurately model spatial and temporal dependencies
plays a crucial role in representing human activities from videos.
For example, an activity performed by a single actor (e.g., walk)
can be characterized a specific spatial organization of body  parts
at each time stamp  (a spatial dependency) as well as by the
evolution of such body parts in time (a temporal dependency); an
activity performed by two (or multiple) individuals (meeting and
talking) can be characterized by a specific arrangement of the
individuals in the scene (a spatial dependency) as well as by the
movement of such individuals in time (a temporal dependency).
Despite recent successes in activity recognition research, finding a
suitable way for capturing both types of temporal and spatial
dependencies still remains an open problem. In this paper we propose
a new unified framework that can be used to simultaneously capture
directional (causal) temporal dependencies and spatial
(bidirectional) dependencies for complex activities recognition.

\begin{figure}
\centering
\includegraphics[width=8.0cm, height=4.0cm]{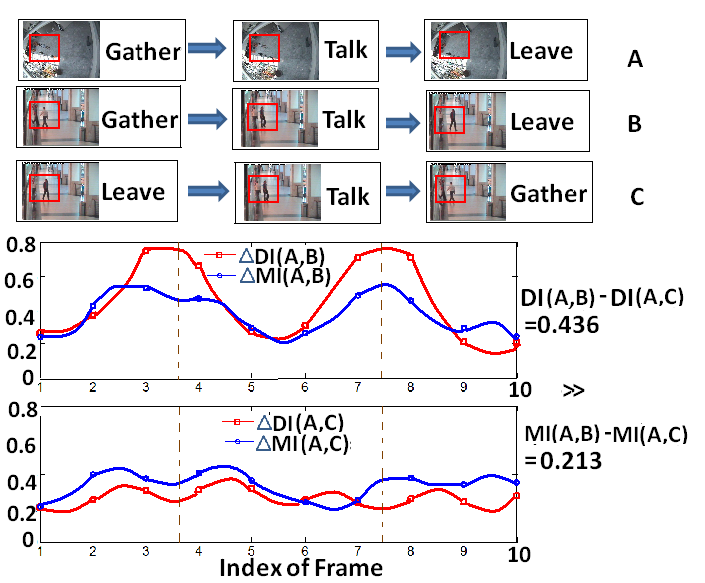}
\caption{Illustration of the advantages of DI (red curves) as
compared to MI (blue curves) for capturing similarities between
causal activities in pairs of video sequences. Videos A and B
contain similar "Gather-talk-leave" event sequences while video C is
a time reversed version of video B. As it is a directional measure,
DI is more sensitive than MI to the ordering of events and this can
be seen from the greater contrast between the two red $\Delta$DI
trajectories in each panel as compared to the lesser contrast
between the two blue $\Delta$MI trajectories($\Delta$DI is the
temporal change of the DI similarity measure over successive frames
and similarly for $\Delta$MI).}\label{MIDI}
\end{figure}

Inspired by a recent contribution in [3], we propose to capture
temporal dependency using the concept of directed information. The
directed information (DI) was first proposed by Massey in 1990 [16]
as an extension of Shannon's mutual information. The DI provides a
decomposition of the MI into its causal and anti-causal components.
Different from MI, DI is a function of the time-aggregated feature
densities extracted from a pair of sequences with strong temporal
structure. In [3], directed information was introduced for
multimodality video indexing. In that work, a Jame-Stein shrinkage
regularization was applied to control the overfitting error called
shrinkage regularized directed information assessment (SODA). SODA
is completely data-driven; it is based on non-parametric estimation
of feature distributions and their associated information
divergences. As shown in Fig.~\ref{MIDI} and described in [3], SODA
is sensitive to the directional information flow for video sequences
across frames. While successful in capturing the ordering of events
when two video sequences are compared, it does not capture spatial
dependencies within each time stamp.

As demonstrated in other approaches for modeling the human pose
[12], a Markov random field formulation (MRF) is suitable for
capturing the structural dependencies among elements in the scene.
Our main idea in this paper is to integrate a MRF approach for
modeling spatial dependencies (among body parts of individuals as
well as among individuals themselves) with a SODA formulation for
modeling the temporal dependencies of individuals in time. Our
SODA+MRF framework combines in a principled fashion a model-free
directed information estimator over time with a model-based
Markovian parts model [4] over space.

\begin{figure}[h]
\begin{center}
\includegraphics[width=8.0cm,height=3.0cm]{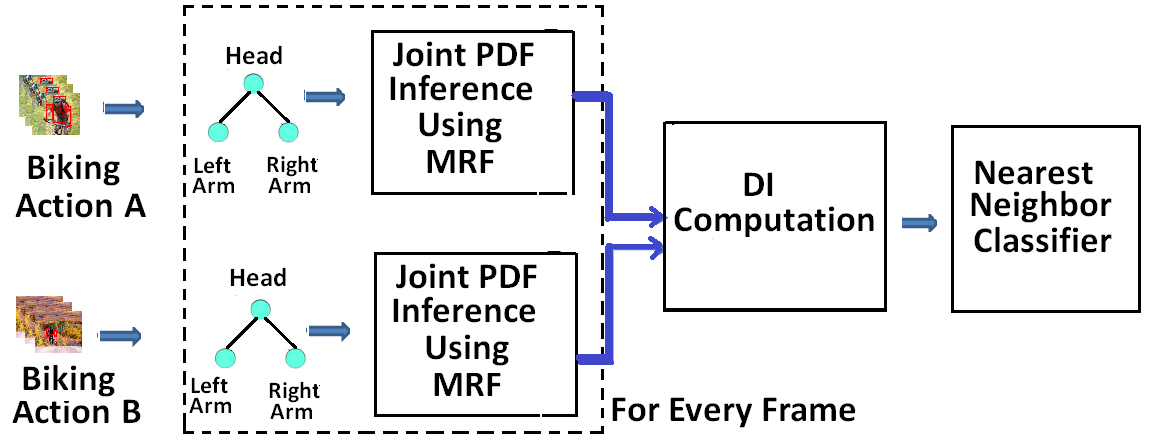}
\end{center}
\vspace{-0.2in} \caption{The proposed pictorial SODA+MRF model
captures dependencies between human body parts and interactions
between people for action recognition. We infer the joint
probability distribution (PDF) of body parts using a MRF for each
video frame. Then, directed information is calculated between pairs
of PDFs across frames and it is used to train a nearest neighbor
classifier for action recognition.}\label{pictorial}
\end{figure}

In the following paragraph we summarize our SODA+MRF approach to
action recognition. We emphasize that the integration of SODA and
MRF is natural and can significantly enhance the detection and
recognition performance in both of the spatial and temporal domains.
As illustrated in Fig.\ref{pictorial}, by detecting the human body
parts and characterizing the human body configurations as a tree
model based on histograms of oriented gradient (HOG) features, the
full joint part distribution in each frame can be inferred using a
MRF. In the presence of multiple people interactions, the spatial
dependency can be inferred using pairwise potentials as function of
their distance and location. Once the joint part distribution in
each frame is estimated, SODA is then used to compute a similarity
measure between pairs of sequences. Finally, a nearest neighbor
classifier is applied to the distances for action classification and
recognition as shown in Fig.\ref{pictorial}.

Our SODA+MRF framework has the following advantages: i) it is robust
to view point changes. Compared to the work by Junejo et al. [10],
that uses self-similarity distance for view invariant recognition,
we take advantage of the information theoretical measure of the flow
of information between past frames to obtain robustness to view
point changes. ii) it yields better or at least comparable
performance in action recognition accuracy compared to several
state-of-the-art approaches including those of Le et al. [21],
Niebles et al. [17], Choi et al. [4] for the UCF sports action
recognition dataset, Olympic dataset and Collective activity
dataset.

\subsection{Related Work}
Action recognition has been extensively investigated in computer
vision. Computing correlations between spatiotemporal (ST) volumes
(i.e., whole video inputs) is the most straightforward method.
Various correlation methods such as cross correlation between
optical flow descriptors [1] and a consistency measure between ST
volumes from their local intensity variations [19] have been
proposed. In the work by Gao \emph{et al.} [9], the authors
discretized the state space and use the loopy belief propagation
algorithm (LBP) for estimating the body pose in 2D for specific
views. In the work by Felzenszwalb \emph{et al.} [7], a
computationally efficient framework for part-based modeling and
recognition of objects was presented. They represented an object by
a collection of parts arranged in a deformable configuration and
used the resulting models to locate the corresponding objects in
novel images. Recently, Ramanan \emph{et al.} [18] developed a
system for tracking the articulations of people from video sequences
by first building an appearance model of each person and then
detecting these models in each frame. In [20], Sundaresan and
Chellappa proposed a general approach using Laplacian Eigenmaps and
a graphical model of the human body to segment 3D voxel data of
humans into different articulated chains. In [14], Laptev \emph{et
al.} utilized automatic collection of realistic samples of human
actions from movies based on movie scripts for automatic learning.
The recognition of complex action classes relies on space-time
interest points and a multi-channel SVM classifier. More recently,
in [10], Junejo \emph{et al.} explored self-similarities of action
sequences over time and developed an action descriptor that captures
the structure of temporal similarities and dissimilarities within an
action sequence. However, these works did not explore the directed
temporal dependency and bidirectional spatial dependency in a
unified framework.

As far as we know, this is the first time that a directed measure
(SODA) coupled with a Markov Random Field (MRF) model has been
proposed for action recognition. The work [15] by Liu and Shah
applied Shannon's mutual information (MI) for selecting high level
visual words for describing and classifying human actions from
videos.
\section{SODA+MRF Frame Work}
Markov random fields (MRF) constitute a simple but powerful means to
model interaction between nearby objects [4]. The basic idea of the
SODA+MRF model is simple: we first infer the joint probability
density distribution (PDF) of the graph connecting body parts within
each frame using a Markov random field (MRF) and subsequently
characterize the temporal dependencies across frames by calculating
directed information between pairs of video sequences based on these
PDFs using SODA. We integrate MRF and SODA into a unified framework
for pictorial structure based action recognition as shown in
Fig.~\ref{pictorial}.
\begin{figure}[h]
\begin{center}
\includegraphics[width=8.0cm,height=4.0cm]{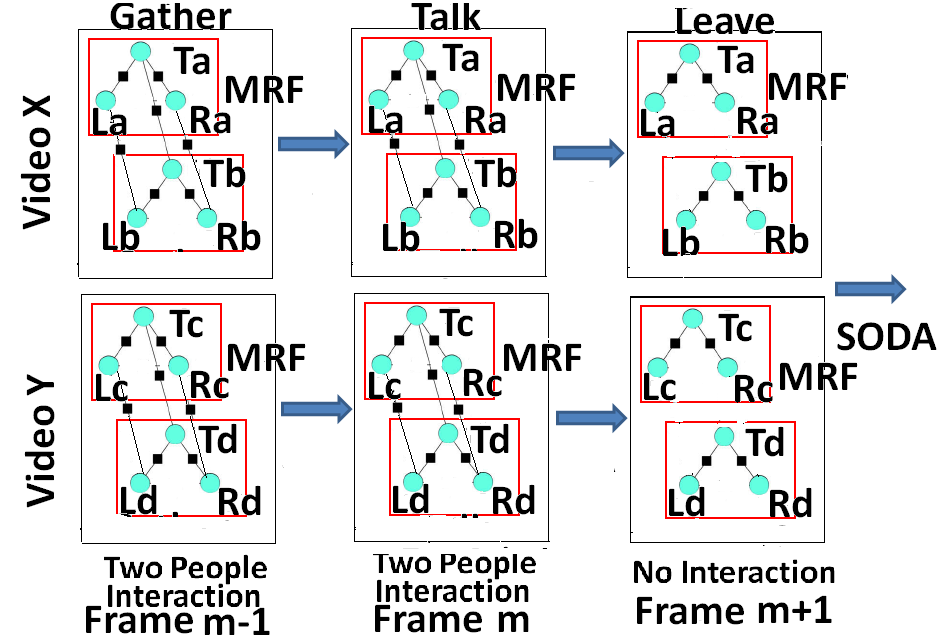}
\end{center}
\caption{The proposed SODA+MRF pictorial structures for two video
sequences for activities "Gather-Talk-Leave" where the proposed
model successfully incorporates the strong interactions between two
people in the activities "Gather" and "Talk" and the weak
interaction between two people in the activity "Leave". Here $T,R,L$
represent the torso, right arm and left arm, respectively, and the
subscripts denote different people and the links characterize the
dependencies. The recognition is achieved by determining whether $X$
(unlabelled) is in the same action category as $Y$
(labelled).}\label{fig:SODAMRF}
\end{figure}
\subsection{MRF for Pictorial Structures}
\textbf{Part-based Understanding with MRF:} In order to compute SODA
between pairs of video sequences, we first estimate the joint
probability density distribution of body parts within each frame
using Markov random field for inference. Taking one person $a$ as an
example, we consider a three-part based model and also a five-part
based model. The three-part based model includes the head and torso,
left arm and right arm denoted by $T_a, L_a, R_a$ respectively. For
the dependencies between different human body parts, as shown in
[18], they can be characterized by a tree- or star-based model as
shown in Fig.\ref{fig:SODAMRF} where only the dependencies between
torsos and the rest of the body parts are considered. In a Markov
random field, given a set of random variables $X^{(m)}=\{T^{(m)}_a,
L^{(m)}_a, R^{(m)}_a\}$, the joint density can be factorized over
the cliques of the graph. For a single person, the joint probability
density distribution of the $m$th frame can therefore be computed by
the product of clique potentials over the maximal cliques
$\textbf{f}(T^{(m)}_a, L^{(m)}_a,
R^{(m)}_a)\propto\psi_{1,1}(T^{(m)}_a,L^{(m)}_a)\psi_{1,2}(T^{(m)}_a,R^{(m)}_a)~,$
where the potential terms $\psi_{1,1}(T^{(m)}_a,L^{(m)}_a)$,
$\psi_{1,2}(T^{(m)}_a,R^{(m)}_a)$ capture the spatial dependencies
between different human body parts, where $\psi_{1,1}(T^{(m)}_a,
L^{(m)}_a)=\exp^{-\gamma_1 d(T^{(m)}_a, L^{(m)}_a)^2}$, where the
$d(T^{(m)}_a, L^{(m)}_a)$ represents the distance between the
centers of the torso and the left arm. The parameter $\gamma_1$
weights the importance of the second type of potentials $\psi_2$
relative to the first potential terms
$\psi_1$.\\
\textbf{Interaction Detection:} We extend our model to multiple
people by considering the dependencies due to human interactions.
Recent work in collective activity recognition [4] suggests that the
interactions between human actions highly depend on the spatial
distance of two people in the video frame and the temporal duration
of the actions. In this work, the spatial interaction between people
is automatically controlled by the pairwise potential term in the
MRF. By considering the interaction between the body parts of two
people, the joint probability density feature distribution of the
$m$th frame can be computed as
\begin{eqnarray}
&&\textbf{f}(X^{(m)})=\textbf{f}(T^{(m)}_a, T^{(m)}_b, L^{(m)}_a,
L^{(m)}_b, R^{(m)}_a,
R^{(m)}_b)\nonumber\\&&\propto\psi_{1,1}(T^{(m)}_a,L^{(m)}_a)\psi_{1,2}(T^{(m)}_a,R^{(m)}_a)\psi_{1,3}(T^{(m)}_b,L^{(m)}_b)\nonumber\\&&\psi_{1,4}(T^{(m)}_b,R^{(m)}_b)\psi_{2,1}(T^{(m)}_a,T^{(m)}_b)\psi_{2,2}(R^{(m)}_a,R^{(m)}_b)\nonumber\\&&\psi_{2,3}(R^{(m)}_a,R^{(m)}_b)~,
\label{MRF}
\end{eqnarray}
where the pairwise potential term  in the $m$th frame is represented
by
$\psi_{2,1}(T^{(m)}_a,T^{(m)}_b)=\exp^{-\gamma_2d(T^{(m)}_a,T^{(m)}_b)^2}$.
Here $d(T^{(m)}_a,T^{(m)}_b)$ represents the distances between the
center of the torsos for two people.
$\psi_{2,2}(R^{(m)}_a,T^{(m)}_b)$ and
$\psi_{2,3}(R^{(m)}_a,T^{(m)}_b)$ are defined similarly. The
potential terms indicates that the interaction increases when the
distance between two people decreases.

The proposed SODA+MRF for two video sequences "Gather-Talk-Leave" is
shown in Fig.\ref{fig:SODAMRF} where both of the strong interaction
between two people in "Gather" and "Talk" and the weak interaction
between two people in "Leave" are incorporated into the model.  Once
the feature vector from the HOG descriptor is obtained, we perform
estimation for the probability density function in MRF model using
Gibbs Sampling (with 500 iterations for burn-in and 1000 iterations
for sampling). The parameters $\gamma_1, \gamma_2$ are estimated
using maximum likelihood. The model can also be easily extended to
the case of more than two people by considering all pair-wise interactions.\\
\textbf{SODA+MRF For Video Sequences:} Once the feature probability
density distribution is estimated from the MRF in each frame, the
temporal dependence is estimated using shrinkage optimized directed
information assessment (SODA). Consider two video sequences $V_x$
and $V_y$ with $M_x$ and $M_y$ frames, respectively. Denote by $X_m$
and $Y_m$ the joint distributions from MRF extracted from the $m$-th
frames of $V_x$ and $V_y$ using MRF, respectively, and define
$X^{(m)}=\{X_k\}_{k=1}^m$ and $Y^{(m)}=\{Y_k\}_{k=1}^m$. The time
aligned directed information (DI) from $V_x$ to $V_y$ is a
non-symmetric generalization of the mutual information (MI) defined
as [16]
\begin{eqnarray}
\DI(V_x\rightarrow V_y)&=&\sum_{m=1}^{M}I(X^{(m)};Y_{m}|Y^{(m-1)})~,
\label{DIformulationp}
\end{eqnarray}
where $M= \min\{M_x,M_y\}$, $I(X^{(m)};Y_{m}|Y^{(m-1)})$ is the
conditional MI between $X^{(m)}$ and $Y_{m}$ given the past
$Y^{(m-1)}$, $I(X^{(m)};Y_{m}|Y^{(m-1)})= E\left[ \ln
\frac{f(X^{(m)},Y_{m}|Y^{(m-1)})}{f(X^{(m)}|Y^{(m-1)})f(Y_{m}|Y^{(m-1)})}\right]$.
An equivalent representation of DI (\ref{DIformulationp}) is in
terms of conditional entropies $ \DI(V_x\rightarrow V_y) =
\sum_{m=1}^{M}\left(H(Y_{m}|Y^{(m-1)})-H(Y_{m}|Y^{(m-1)},X^{(m)})\right),
$ which gives the intuition that the DI is the cumulative reduction
in uncertainty of frame $Y_m$ when the past frames $Y{(m-1)}$ of
$V_y$ are supplemented by information about the past and present
frames $X^{(m)}$ of $V_x$. The DI prescribes a decomposition of the
MI into a sum of causal and anti-causal DIs. It is for this reason
that the MI masks directional information revealed by the DI. The
SODA approach gives a DI estimator that is specifically adapted to
the high dimension of video sources.

To apply SODA we first implement scalar quantization of the
part-based PDF in each video frame. For a single frame the
quantization has an alphabet of $p$ symbols $\mathcal
X=\{x_i\}_{i=1}^p$ corresponding to $p$ quantization cells (classes)
$\mathcal C = \{C_i\}_{i=1}^p$. For a particular frame sequence
$X^{(m)}$ let there be $n$ feature realizations and let
$Z=[z_1,\ldots,z_{p^m}]$ denote the histogram of these realizations
over the respective quantization cells.  Then if the feature
realizations are i.i.d., $Z$ is multinomially distributed with
probability mass function
$P_\theta(z_1=n_1,\ldots,z_{p^m}=n_{p^m})=\frac{n!}{\prod_{k=1}^{p^m}n_{k}!}\prod_{k=1}^{p^m}\theta_{k}^{n_k},$
where $\theta=E[Z]/n=[\theta_1,\ldots,\theta_{p^m}]$ is a vector of
class probabilities and $\sum_{k=1}^{p^m}n_k=n$,
$\sum_{n=1}^{p^m}\theta_k=1$.

In order to implement the DI, the vector of multinomial parameters
$\theta$ must be empirically estimated from the video sequences.
However, since the feature dimension is large the number of unknown
parameters greatly exceeds the number of feature instances.
Substitution of maximum likelihood (ML) estimates in place of
$\theta$, will produce severe overfitting errors. SODA applies
James-Stein shrinkage approach to improve the MSE of the plug-in
estimator, as described in [3]. This approach is based on shrinking
the ML estimator of $\theta$ towards a target distribution $t$ with
uniform distribution, $\hat{\theta}_{k}^{\lambda}=\lambda
t_{k}+(1-\lambda)\hat{\theta}_{k}^{ML}~,$ where $\lambda\in [0,1]$
is a shrinkage coefficient used to optimize DI estimation
performance. The James-Stein plug-in entropy estimator is defined
as:
$\hat{H}_{\hat{\theta}^\lambda}(X)=-n\sum_{k=1}^{p}\hat{\theta}_{k}^{\lambda}\log(\hat{\theta}_{k}^{\lambda})$.
The corresponding SODA plug-in estimator for DI is simply
$\widehat{\DI}^{\lambda}=\DI_{\hat{\theta}^\lambda}(V_x\rightarrow
V_y)$. The optimal value of $\lambda$ that minimizes estimator MSE
is [3]: $\lambda^{\circ}=\arg\min_{\lambda}
E(\widehat{DI}^{\lambda}-DI)^{2}~\label{cost}.$ The resultant
shrinkage optimized DI estimator,
$\widehat{DI}^{\lambda_{\circ}}(X^{M}\rightarrow Y^{M})$, is a
James-Stein DI estimate that yields the SODA estimate.
\subsection{SODA+MRF-based pictorial structure recognition algorithm}
SODA can be used for temporal localization of interactions by
aligning the video sequences. Once the DI optimal shrinkage
parameter has been determined, the local DI is defined similarly to
the DI except that, for a pair of video sequences $X$ and $Y$, the
signals are time shifted and windowed prior to DI computation.
Specifically, let $\tau_x\in [0,M_x-T]$, $ \tau_y\in [0,M_y-T]$ be
the respective time shift parameters, where $T \ll \min\{M_x ,M_y\}$
is the sliding window width, and denote  by $X_{\tau_x}^{M_x}$,
$Y_{\tau_y}^{M_y}$ the time shifted sequences. Then the local DI,
$\mathrm{DI}(X^{M_x}_{\tau_x} \rightarrow Y^{M_y}_{\tau_y})$,
computed using (\ref{DIformulationp}), defines a surface over
$\tau_x,\tau_y$. We use the peaks of the local DI surface to detect
and localize the interactions in the pair of video sequences. We
summarize the algorithm below:
\begin{enumerate}
\item Detect and localize the articulated human body parts for each
video frame using the object detection algorithm [8]. Extract HOG
features from each detected human regions. Compute the joint
distribution of the graph for each frame according to the graph
structure using a MRF by equation (\ref{MRF}).
\item Use SODA to calculate local DI for the pair of video sequences and use the optimal sliding
window learned from the training phase to detect the local peaks in
the DI surface. We define the local DI between video pairs $X$ and
$Y$ as $DI(Y_i, X_j)$ where $i,j$ is the time index in the video
sequence. We generate a quantitative measure of the statistical
significance of each p-value using the expression $1-
\Phi\left(\frac{\hat{D}_{ij}-\mu_{ij}}{\sigma_{ij}}\right)$ on these
DI estimates where $\mu_{ij}$ and $\sigma_{ij}$ stands for the mean
and the variance, $\Phi$ here stands for the cumulative distribution
function of Gaussian distribution. The p-values correspond to the
probability that the observed DI peak values follow the null
hypothesis that the mean DI is equal to zero. The test statistic is
computed as $T=DI(Y,X)=\max_{i,j} DI(Y_i,X_j), (i,j\in
\mathbb{Z}^{+})$ Threshold the DI and $p$-value matrices to find
common activities exhibiting large and statistically significant DI.
\item Test the $K \times (K-1)$ hypotheses that there is a
significant interaction (both directions) between pairs of video
frames containing $K$ frames. Since there are $K(K-1)$ different DI
pairs of video frames, this is a multiple hypothesis testing problem
and we control false discovery rates (FDR) using the corrected
Benjamini- Hochberg (BH) procedure [2]. Finally, a nearest neighbor
classifier is applied on pair-wise DI for classification and
recognition.
\end{enumerate}
\section{Experimental Results}
\textbf{Dataset used:} We evaluate the performance of SODA+MRF model
on three datasets: the UCF50 action recognition dataset, the Olympic
Sports Dataset and the Collective Activity Dataset. (i) UCF50 is an
action recognition dataset with 50 annotated action categories,
consisting of realistic videos taken from YouTube including:
Baseball Pitch, Basketball Shooting, Bench Press, Biking, Billiards
Shot, Breaststroke, Clean and Jerk, Diving, Drumming, Fencing, Golf
Swing, Playing Guitar. For all the 50 categories, the videos are
grouped into 25 groups, where each group consists of more than 4
action clips. The video clips in the same group may share some
common features, such as the same person, similar background,
similar viewpoint, and so on. (ii) The Olympic Sports Dataset
contains videos of athletes practicing different sports with 16
sport classes and 50 sequences per class. (iii) The collective
activity dataset contains 7 different collective activities:
crossing, walking, waiting, talking, queueing, jogging and dancing
and 60 video sequences with varying view point.

Whenever we report performance comparisons in the following
experiments, a 5 part model is utilized. The five-part based model
includes heads and torsos, left arms, right arms, left legs and
right legs. The pairwise DI was computed using the aforementioned
SODA+MRF method for all pairs of video clips in the database. The
pairwise symmetrized DI's were then used to train a nearest neighbor
classifier where the symmetrized DI is the sum of DI from the video
sequence $X$ to $Y$ and $Y$ to $X$. Half of the videos were randomly
selected for training
and the remainder were used for testing.\\
\textbf{Feature Representation:} Our model of human actions can be
applied over a variety of video descriptors. In this paper, the
human body parts are detected using the method presented in [8]
which is a complete learning-based system for detecting and
localizing objects in images. The features within the detection
windows for body part configurations are represented with histogram
of oriented gradient (HOG) descriptors. The histogram of oriented
gradient (HOG) descriptors [5] contain four main steps including
gamma/color normalization, gradient computation, spatial/orientation
binning and normalization and descriptor blocks. In practice this is
implemented by dividing the image window into small spatial regions
called cells, for each cell accumulating a local 1-D histogram of
gradient directions or edge orientations over the pixels of the
cell. The number of orientation bins is selected to be 9 bins spaced
over 0$^\circ$-180$^\circ$. We utilize
3$\times$3 cell blocks of 6$\times$6 pixel cells for detection where detectors are based on rectangular bounding boxes. \\
\textbf{Competing Algorithm investigated:} For the UCF action
recognition dataset, the performance will be compared to several
state-of-the-art approaches and baseline methods including: Le
\emph{et al.}[21], Wang \emph{et al.} [22], Laptev \emph{et al.}
[14][13] and Klaser \emph{et al.} [11]. We also compare to SODA
without MRF [3] as a baseline method. In [21], Le \emph{et al.}
presented an extension of the Independent Subspace Analysis
algorithm to learn invariant spatio-temporal features from unlabeled
video data. The performance on Olympic sports dataset is compared to
the method by Niebles \emph{et al.} [17] and [14]. In [17], the
authors presented a framework for modeling motion by exploiting the
temporal structure of human activities. The performance on the
collective activity dataset is compared to Spatial Temporal Volume
representation (STV) and Randomized Spatial Temporal volume
representation (RSTV) [4] where RSTV is constructed on a Random
Forest structure which randomly samples variable volume
spatial-temporal regions to pick the most discriminating attributes
for classification.
\subsection{Evaluation with UCF action recognition dataset}
\begin{figure}[h]
\begin{center}
\includegraphics[width=8.0cm,height=4.0cm]{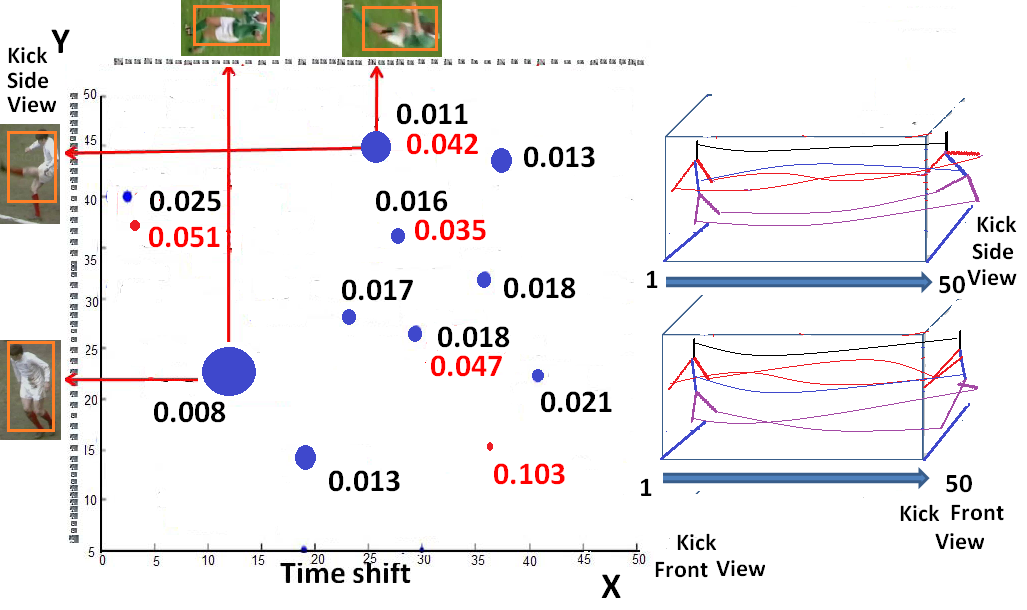}
\end{center}
\vspace{-0.2in}\caption{Bubble graph of peak values for local DI in
$\widehat{\mathrm{DI}}(X^{M_x}_{\tau_x}\rightarrow
Y^{M_y}_{\tau_y})$ between pairs of UCF videos using the same class
of the activity from different views: $X$ (kick-front-view) and $Y$
(kick-side-view) using SODA+MRF model (black) and SODA model (red).
Here the axes range over $\tau_x$ and $\tau_y$, which denote time
shift parameters of the respective video frames, and the sliding
window width is $T=7$ frames. The 10 most significant theoretical
p-values (blue circles whose sizes are proportion to statistical
significance of the DI peaks) are shown to indicate the most
statistically reliable DI peaks. The SODA+MRF model achieves more
significant statistical p-values compared to SODA only. The results
indicate that the most similar two actions happen at the 12th and
28th frames in $X$ and the 22th and 43th frames in $Y$. The size of
the bubble is inverse proportional to the
p-values.}\label{fig:SODAMRF1}
\end{figure}

\textbf{Localization of Common Activities:} In order to gain more
insight into the proposed method for part-based understanding, we
show that SODA+MRF is capable of accurately recognizing single human
actions independent of viewpoint. In Fig.\ref{fig:SODAMRF1}, we
illustrate the local DI between a video pair, denote from $X$ to $Y$
was rendered as a surface over $\tau_x,\tau_y$ for a FDR of 0.1
where the FDR is the expected proportion of false positives among
all significant hypotheses, as explained in Section 2.2. The peaks
on this surface were used to detect and localize common activities,
i.e., activities in $X$ that were predictive of activities in $Y$.
SODA+MRF detects 10 interactions while SODA without MRF only detects
5 interactions. The bubbles (dots) in Fig.\ref{fig:SODAMRF1} occur
at the peaks of the pairwise DI using the window size $T=7$ with the
highest average precision and the size of each bubble inversely
proportional to the corresponding p-value. The figure shows that the
most similar actions "kick" occur at the 12th and 28th frames in X
(front view) and the 22th and 43th frames in Y (side view) denoted
by DI peaks. Since directed information calculates the accumulated
information for the current frame given all of the past frames for
the purposeful action "kick", although the appearances from two
sequences have variations due to different views, the intrinsic
distance based on DI is robust to the change of view. The
performance continues to be significant over a range of practical
FDR thresholds (0.1 to 0.05). Compared to the recognition result
using SODA only \cite{TMM}, the SODA+MRF method significantly
improves the recognition performance, as indicated by the more
highly concentrated  peaks of the inverse p-values in
Fig.\ref{fig:SODAMRF1}. Since both SODA+MRF and SODA utilize
directed information to capture the temporal structures, the
superior performance can be mainly attributed to the use of the
pictorial structures by coupling the Markov random field with SODA
to model the spatial dependency between different human body
configurations and interactions.

\begin{figure}[h]
\begin{center}
\includegraphics[width=8.0cm,height=4.0cm]{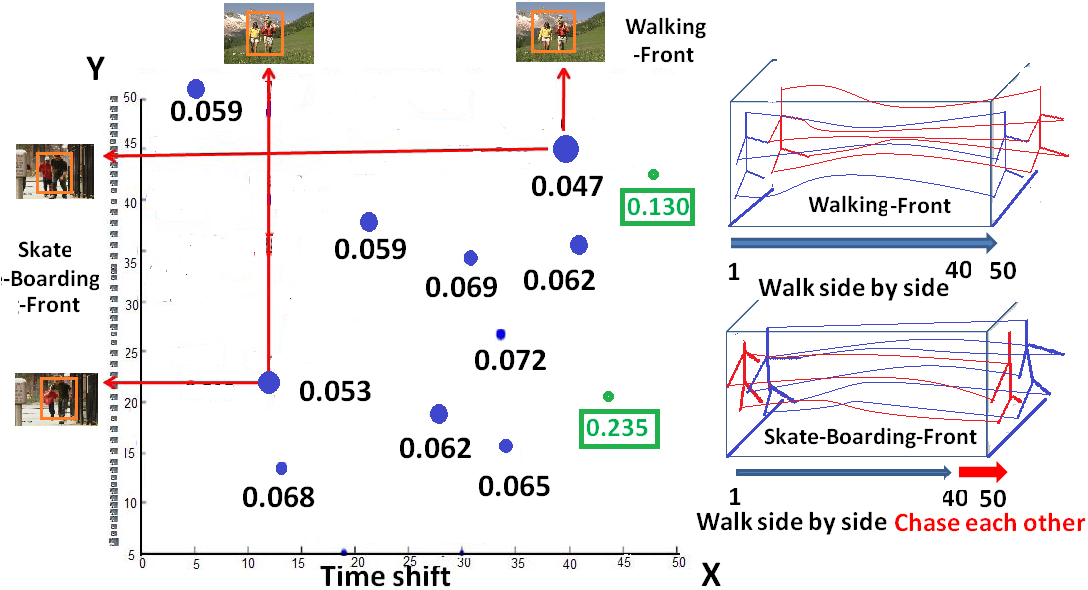}
\end{center}
\vspace{-0.2in} \caption{Bubble graph of peak values for local DI in
$\widehat{\mathrm{DI}}(X^{M_x}_{\tau_x}\rightarrow
Y^{M_y}_{\tau_y})$ between UCF videos using different activities:
$X$ (walk-front-view) and $Y$ (skate-boarding-front-view) with
similar setting as Fig.\ref{fig:SODAMRF1} where in the 1st to 40th
frames, the two sequences contain similar actions "walking side by
side" and in the 41st to 50th frames, they have dissimilar actions
with "walking" and "chasing" respectively. The top 10 true positives
discovered by our methods are annotated with the corresponding
p-values in black. The detections of false positives from SODA are
highlighted with green bounding boxes. The results indicate that the
most similar two actions happen at the 12th and 40th frames in $X$
and the 23th and 44th frames in $Y$.}\label{fig:SODAMRF2}
\end{figure}

\textbf{Interaction Detection:} Moreover, we demonstrate that
SODA+MRF can successfully discriminate different actions for
multiple people with interaction. In these two sequences
"walk-front" and "skate-boarding-front", from the 1st to the 40th
frames, the actions are similar since both of them contain two
people walking side by side. From the 41st to 50th frames, in the
sequence of "skate-boarding", two people start to chase each other
where the action "chasing" is characterized with much stronger
temporal directional dependency and larger spatial distances
compared to the action "walk side by side". As shown in
Fig.\ref{fig:SODAMRF2}, our scheme is sensitive in detecting such
interaction changes by modelling the spatial and temporal
dependencies accurately. The results indicate that the most similar
two actions happen at the 12th and 40th frames in $X$ and the 23th
and 44th frames in $Y$. All of the 10 true positives are detected in
the first 40 frames, where the positions of the true positives
represents the frame indices for the most similar frames in two
sequences. Compared to the use of the SODA method only, our SODA+MRF
method also successfully removed two false positives (highlighted in
green in Fig.\ref{fig:SODAMRF2}). Again, the advantage of our method
is attributable to the coupling of MRF with SODA that allows it to
capture complex interactions between people with significant
background variation.

\textbf{Comparisons:} We first report the average accuracy of
several baseline methods for recognition in order to demonstrate the
superiority of the proposed method, as shown in Table 1. The
baseline methods include: MRF+MI, 3D MRF and SODA+MRF without
interaction where the MRF+MI method models the spatial dependency
using a MRF and the temporal dependency using mutual information;
the 3D MRF models both the spatial and temporal dependencies using a
three dimensional MRF. The SODA+MRF model without interaction
ignores all the interaction between people by setting the pairwise
potential $\psi_2$ to be a constant. A thorough comparison of the
mean average precision for UCF action recognition dataset with the
state-of-the-art approaches is shown in Table 1. It can be seen that
our SODA+MRF method not only achieves significant advantages over
the three baseline methods but also outperforms a wide range of the
state-of-the-art approaches. Furthermore, the table indicates that
by considering the interaction between different human body
configurations, SODA+MRF improves the recognition accuracy by 4.8\%.
Compared to the best published performance of [21] (86.5\%), our
SODA+MRF method has an improved mean average precision of 88.3\% for
the UCF action recognition dataset. We also compare computational
complexity in Table 2 for a machine running Matlab with 2.26GHz CPU
and 24Gb RAM.

\begin{table}
\begin{tabular}{|c|c|}
  \hline
  Algorithm & Mean AP \\
  \hline
  Harris 3D [13] + HOG/HOF [14](from [22]) & 78.1\% \\
  \hline
  Cuboids [6] + HOG3D [14] (from [22]) & 82.9\% \\
  \hline
  Hessian [23] + HOG/HOF (from [22]) & 79.3\% \\
  \hline
  Hessian [23] + ESURF [23] (from [22]) & 77.3\% \\
  \hline
  Dense + HOF [14] (from [22]) & 82.6\% \\
  \hline
  Dense + HOG3D [11] (from [22]) & 85.6\% \\
  \hline
  Hierarchical Spatial Temporal Feature [21] & 86.5\% \\
  \hline
  MRF+MI (baseline) & 75.2\% \\
  \hline
  3D MRF (baseline) & 84.1\% \\
  \hline
  MRF (without interaction)+SODA & 83.5 \% \\
  \hline
  SODA+HOG [3] & 84.7\% \\
  \hline
  Our method (SODA+MRF) & \textbf{88.3\%} \\
  \hline
\end{tabular}
\caption{Average Accuracy Comparison for UCF Sports Action Dataset.}
\end{table}

\begin{table}
\begin{tabular}{|c|c|}
  \hline
  Algorithm & Seconds/Frame \\
  \hline
  HOG3D [11] & 0.22 \\
  \hline
  Hierarchical ST Feature (2 layer) [21] & 0.44 \\
  \hline
  SODA+MRF & 0.41 \\
  \hline
\end{tabular}
\caption{Computational time comparison.}
\end{table}
\subsection{Evaluation on Olympic sports dataset}
Here we evaluate SODA+MRF on the Olympic sports dataset which
includes strong temporal structures. We demonstrate improvements in
average precision for the classification task in Table 3, where our
method achieves average precision of 75.1\% for all the 16
activities compared to 72.1\% (Niebles et al. [17]) and 62.0\%
(Laptev et al.[14]). Our method achieves the best average precisions
for 12 of the 16 activities. These results indicate the efficacy of
our SODA+MRF method in recognizing complex human actions.
\subsection{Evaluation on collective activity dataset}
\begin{figure}[h]
\begin{center}
\includegraphics[width=8.0cm,height=4.0cm]{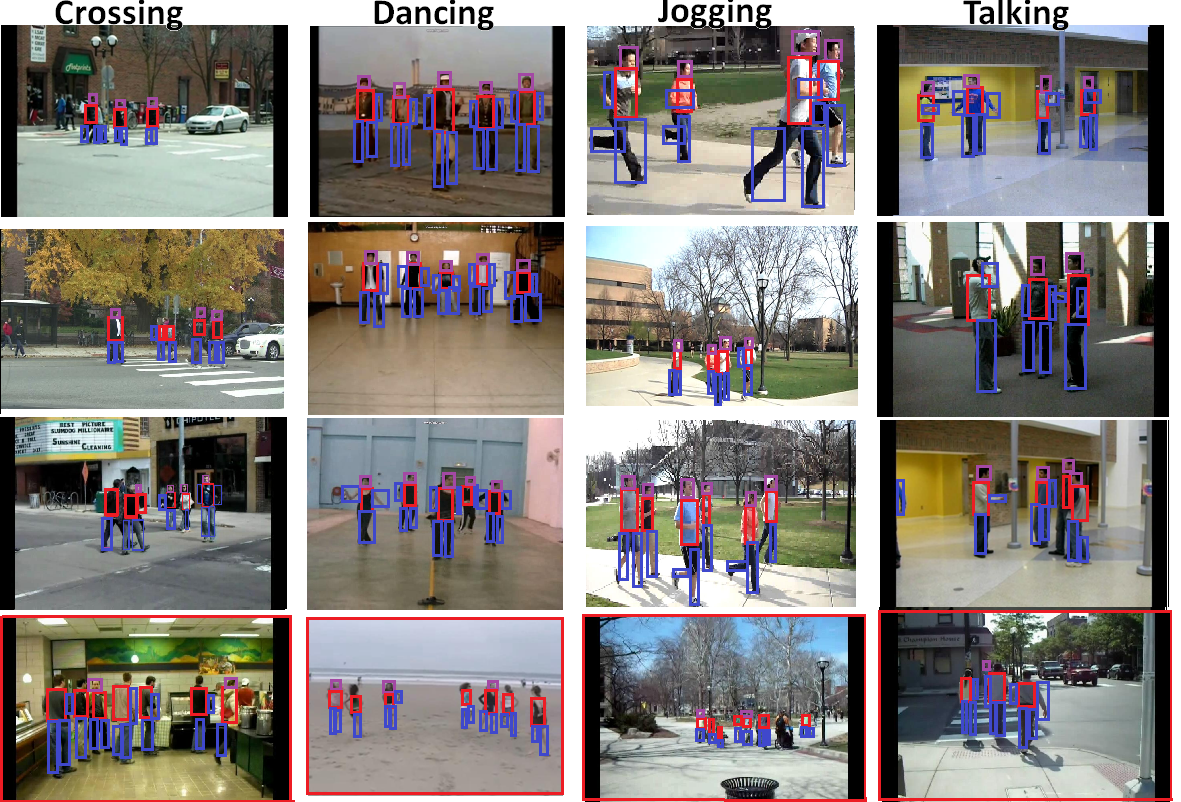}
\end{center}
\caption{Example results on the 6-category dataset. Top 3 rows show
examples of good classification and bottom row shows examples of
false classification with SODA+MRF model, where the color encodings
are as follows: head (purple), torso (red), arms and legs (blue).
}\label{collective}
\end{figure}
\begin{table*}
\begin{tabular}{|c|c|c|c|c|c|c|c|}
  \hline
  Sport & Niebles [17] & Laptev [14] & ours & Sport & Niebles [17] & Laptev [14] & ours \\
  \hline
  high-jump & 68.9\% & 58.4\% & \textbf{71.3\%} & javelin-throw & \textbf{74.6\%} & 61.1\% & 72.7\% \\
  \hline
  long-jump & 74.8\% & 66.8\% & \textbf{77.9\%} & hammer-throw & 77.5\% & 65.1\% & \textbf{79.5\%} \\
  \hline
  triple-jump & 52.3\% & 36.1\% & \textbf{58.3\%} & discuss-throw & 58.5\% & 37.4\% & \textbf{63.2\%} \\
  \hline
  pole-vault & \textbf{82.0\%} & 47.8\% & 81.6\% & diving-platform & 87.2\% & \textbf{91.5\%} & 89.9\% \\
  \hline
  gymnastic-vault & 86.1\% & \textbf{88.6\%} & 87.4\% & diving-springboard & 77.2\% & 80.7\% & \textbf{84.2\%} \\
  \hline
  short-put & 62.1\% & 56.2\% & \textbf{65.7\%} & basketball-layup & 77.9\% & 75.8\% & \textbf{79.1\%} \\
  \hline
  snatch & 69.2\% & 41.8\% & \textbf{73.2\%} & bowling & 72.7\% & 66.7\% & \textbf{75.6\%} \\
  \hline
  clean-jerk & 84.1\% & 83.2\% & \textbf{86.2\%} & tennis-serve & 49.1\% & 39.6\% & \textbf{55.2\%} \\
  \hline
\end{tabular}
\caption{Average Precision for classification task on Olympic sports
dataset, where our method achieves the average precision for all the
16 activities of 75.1\% compared to 72.1\% (Niebles et al.[17] ) and
62.0\% (Laptev et al.[14]).}
\end{table*}

We show results on the 6-category collective activity dataset in
Fig.\ref{collective}. In Table 3, we demonstrate final
classification accuracy for the 6-category dataset by comparing
SODA+MRF with the RSTV method [4] which previously had the best
reported performance for this dataset for activities of crossing,
waiting, queueing, talking, dancing and jogging. As shown in Table
4, the SODA+MRF method achieves an average precision 0.847, which is
slightly better than RSTV having average precision 0.82. However, if
one excludes the "jog" activity for which RSTV is the top performer,
SODA+MRF performs uniformly better than RSTV, achieving average
precision of 0.846 (SODA-MRF) as compared with 0.796 (RSTV) overall.
This demonstrates that the proposed method is effective in capturing
the subtle spatial and temporal interactions between body
configurations compared to RSTV which detects general interactions
between people. We believe that SODA+MRF underperforms for the
jogging activity is mainly due to its difficulty in detecting
non-frontal faces and occlusions. A refined implementation of the
MRF parts model that accounts for occlusions may overcome this
difficulty. This is future work.
\begin{table*}[!htb]
\begin{tabular}{|c|c|c|c|c|c|c|}
  \hline
  SODA+MRF/RSTV[4] & cross & wait & queue & talk & dance & jog \\
  \hline
  cross & \textbf{82.9}/76.5\% & 4.2/6.3\% & 1.6/1.6\% & 0/0\% & 0/0\% & 11.3/15.6\% \\
  \hline
  wait & 2.7/4.8\% & \textbf{84.3}/78.5\% & 7.6/12.8\% & 2.3/0.9\% & 1.6/3.1\% & 1.5/0\% \\
  \hline
  queue & 0/0.2\% & 12.5/20.1\% & \textbf{83.6}/78.5\% & 2.3/0.8\% & 1.0/0.4\% & 0.6/0\% \\
  \hline
  talk & 2.3/2.8\% & 4.8/6.1\% & 4.2/6.5\% & \textbf{87.2}/84.1\% & 1.1/0.5\% & 0.4/0\% \\
  \hline
  dance & 6.3/11.1\% & 3.9/5.1\% & 2.6/2.9\% & 1.3/0.4\% & \textbf{85.1}/80.5\% & 0.8/0.1\% \\
  \hline
  jog & 7.1/5.9\% & 0/0 & 0.4/0 & 0/0 & 0/0 & 92.5\textbf{/94.1\%} \\
  \hline
\end{tabular}
\caption{Final classification accuracy for 6-category dataset for
the activities crossing, dancing, jogging and talking, where our
proposed SODA+MRF method achieves average precision \textbf{0.86}
which is better than RSTV \textbf{0.82}.}
\end{table*}

\section{Conclusion}
We proposed a novel framework called SODA+MRF for activity
recognition. The spatial dependency between human body
configurations and the interaction of these configurations are
characterized by a Markov Random Field model. SODA+MRF combines the
MRF model with a James-Stein shrinkage approach to DI estimation,
resulting in minimum mean squared error to capture the temporal
dependency across frames. We illustrated the SODA+MRF model for
activity detection/localization with the UCF action recognition, the
Olympic sports dataset and the collective activity databases. Our
results indicate that the SODA+MRF model is able to discriminate
video events that involve strong human interactions and demonstrates
better action recognition performance as compared to several
state-of-the-art approaches.



\section*{Reference}
{[1] A.~A.Efros, A.~C.Berg, G.~Mori and J.~Malik,  {\it Recognizing
Actions at a Distance}, IEEE International Conference on Computer
Vision, 2003.

[2] Y. Benjamini and D. Yekutieli {\it The control of the false dis-
rate in multiple testing under dependency.}. The Annals of
Statistics, 2001.

[3] X. Chen, A. Hero, and S. Savarese, {\it Multimodality Video
Indexing and Retrieval Using Directed Information}, IEEE
Transactions on Multimedia, 2011.

[4] W. Choi, K. Shahid, and S. Savarese, {\it Learning Context for
Collective Activity Recognition}, In IEEE Conference on Computer
Vision and Pattern Recognition. IEEE, 2011.

[5] N. Dalal and B. Triggs, {\it Histograms of Oriented Gradients
for Human Detection}, IEEE Conference on Computer Vision and Pattern
Recognition, 2005.

[6] P. Dollar, V. Rabaud, G. Cottrell, and S. Belongie., {\it
Behavior recognition via sparse spatio-temporal features.}, ICCV
workshop: VS-PETS, 2005.

[7] P. Felzenszwalb and D. Huttenlocher, {\it Pictorial Structures
for Object Recognition.}, In International Journal of Computer
Vision. IEEE, 2005.

[8] P. Felzenszwalb, D. McAllester, and D. Ramaman {\it A
Discriminatively Trained, Multiscale, Deformable Part Model.}, in
IEEE CVPR, 2008.

[9] J. Gao and J. Shi, {\it Multiple frame motion inference using
belief propagation}, In IEEE International Conference on Automatic
Face and Gesture Recognition. IEEE, 2004.

[10] I. Junejo, E. Dexter, and P. Laptev, I.and Perez, {\it View-
Independent Action Recognition from Temporal Self- Similarities}, in
IEEE Transactions on Pattern Analysis and Machine Intelligence, Vol.
33, IEEE 2011.

[11] A. Klaser, M. Marszalek, and C. Schmid, {\it A spatial-temporal
descriptor based on 3D descriptor.}, In Bristish Machine Vision
Conference. IEEE, 2008.

[12] P. Kohli, J. Rihan, M. Bray, and P. Torr., {\it Simultaneous
segmentation and pose estimation of humans using dynamic graph
cuts}, In International Journal of Computer Vision, 2008.

[13] I. Laptev and T. Linderberg, {\it Space-time Interest Points},
International Conference on Computer Vision, 2003.

[14] I. Laptev, M. Marszalek, C. Schmid, and B. Rozenfeld, {\it
Learning realistic human actions from movies}, In IEEE Conference on
Computer Vision and Pattern Recognition, 2008.

[15] J. Liu and M. Shah {\it Learning human actions via information
maximization}. CVPR, 2008.

[16] J.~Massey. {\it Causality, feedback and directed information.
Symp Information Theory and Its Applications (ISITA)}, 1990.

[17] J. C. Niebles, C.W. Chen, and L. Fei-Fei., {\it Modeling
Temporal Structure of Decomposable Motion Segments for Activity
Classification}, IEEE ECCV 2010.

[18] D. Ramanan, D. Forsyth, and A. Zisserman, {\it Tracking People
by Learning Their Appearance}, In Pattern Analysis and Machine
Intelligence. IEEE, 2007.

[19] E. Shechtman and M. Irani., {\it Space-Time Behavioral
Correlation}, In IEEE Conference on Computer Vision and Pattern
Recognition (CVPR). IEEE, 2005.

[20] A. Sundaresan and R. Chellappa, {\it and R. Chellappa. Model
Driven Segmen- of Articulating Humans in Laplacian Eigenspace.}, In
IEEE Transactions on Pattern Analysis and Machine Intelligence,
volume 30. IEEE, 2008.

[21] Q. V.Le, W. Y.Zou, S. Y.Yeung, and A. Y.Ng., {\it Learning
hierarchical spatio-temporal features for action recognition with
independent subspace analysis}, In IEEE Conference on Computer
Vision and Pattern Recognition. IEEE, 2011.

[22] H.Wang, M. M.Ullah, A. Klaser, and C. Schmid., {\it Evaluation
of local spatial-temporal features for action recognition}, In
Bristish Machine Vision Conference. IEEE, 2010.

[23] G.Willems, T. Tuytelaars, and L. V.Gool, {\it An efficient
dense and scale-invariant spatial-temporal interest point
detector.}, In European Conference on Computer Vision. IEEE, 2008. 






\end{document}